\documentclass{article} 
\usepackage{iclr2027_conference,times}

\usepackage{amsmath,amsfonts,bm}

\def\eqref#1{equation~\ref{#1}}

\def\1{\bm{1}}

\DeclareMathAlphabet{\mathsfit}{\encodingdefault}{\sfdefault}{m}{sl}
\SetMathAlphabet{\mathsfit}{bold}{\encodingdefault}{\sfdefault}{bx}{n}

\usepackage{hyperref}
\usepackage{url}
\usepackage{graphicx}
\usepackage{booktabs}
\usepackage{multirow}
\usepackage{natbib}
\usepackage{wrapfig}

\title{SPACE: Sparse Predictive Attractor via Counterfactual Eviction for Streaming Video Memory}

\author{
\begin{tabular}{@{}l@{}}
\bfseries
Hongjin Niu$^{1}$ \quad
Weizhan Zhang$^{1}$\thanks{Corresponding author.} \quad
Shuo Bao$^{2}$ \quad
Jiahao Wang$^{1}$ \quad
Muyan Jiao$^{1}$
\\[-1pt]
\bfseries
Kairui Wen$^{3}$ \quad
Yong-Jin Liu$^{3}$
\end{tabular}
\\[6pt]
\normalfont\small
$^{1}$School of Computer Science and Technology, Xi'an Jiaotong University
\\[-1pt]
\normalfont\small
$^{2}$School of Advanced Manufacturing and Robotics, Peking University
\\[-1pt]
\normalfont\small
$^{3}$Department of Computer Science and Technology, Tsinghua University
\\[3pt]
\normalfont\small
\texttt{niuhongjin@stu.xjtu.edu.cn}
\\[-1pt]
\normalfont\small
\texttt{zhangwzh@xjtu.edu.cn}
}
\iclrfinalcopy 
\begin{document}

\maketitle
\maketitle

\fancyhead{}
\renewcommand{\headrulewidth}{0pt}

\begingroup
\renewcommand{\thefootnote}{}
\footnotetext{\textbf{Code:} \url{https://github.com/GilliaN-HJ/SPACE}}
\endgroup

\begin{figure}[h]
    \centering
    \includegraphics[width=1\linewidth]{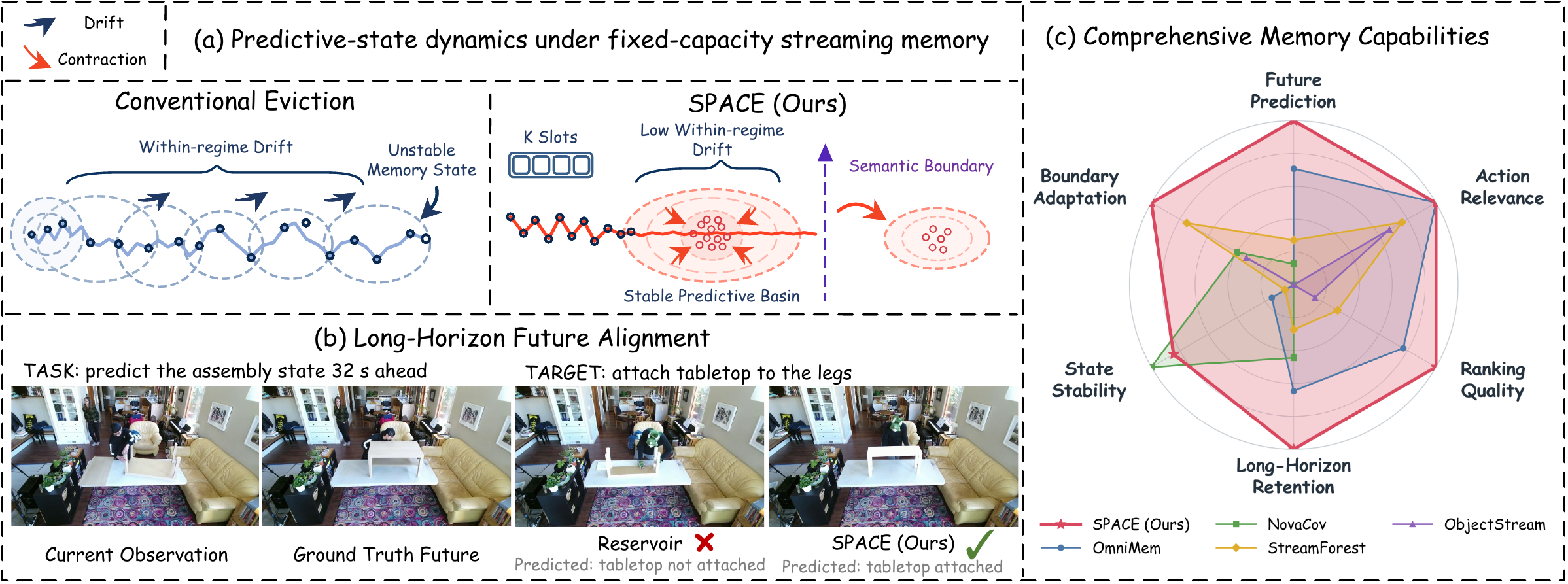}
    \caption{\textbf{Predictive stability of SPACE.} Unlike conventional
    streaming memory, which drifts under repeated updates, SPACE forms a stable
    yet adaptive predictive attractor, improving long-horizon future alignment
    while preserving broad memory capabilities. In (c), the six metrics are independently min–max normalized to \([0,1]\) across all methods, with directions aligned so that higher is better; radial rings are spaced by 0.2. Predicted futures are
    nearest-neighbor RGB visualizations of future latents from a
    participant-disjoint RGB retrieval bank, not generated pixels.}
    \label{fig:teaser}
\end{figure}

\begin{abstract}

Fixed-capacity streaming video memory requires repeated eviction decisions
whose effects accumulate over time. Yet existing policies are evaluated
primarily in terms of retained information or downstream accuracy, leaving
how repeated updates alter the futures supported by memory largely
unexamined. We define a memory's predictive state as the future
representations supported by its retained history and formulate eviction as
counterfactual control over transitions in this space. We introduce
\textbf{SPACE} (Sparse Predictive Attractor via Counterfactual Eviction),
which uses a frozen multi-horizon JEPA to predict the future representations
induced by alternative eviction actions. Counterfactual utility identifies
future-useful alternatives, while slow predictive-basin geometry determines
when to correct avoidable drift and when to adapt to sustained predictive
change, without online parameter updates. We further introduce
\textbf{MABS-Bench}, which evaluates future-task sufficiency, within-regime
predictive stability, transition responsiveness, and perturbation recovery
under matched causal streams and memory budgets. Across multiple video
datasets, SPACE yields consistent improvements in dataset-native task
performance while reducing predictive-state drift.

\end{abstract}

\section{Introduction}

Streaming video models must summarize an indefinitely growing observation
stream using a memory of fixed capacity
~\citep{xu2021long,wu2022memvit,he2024ma,qian2024streaming,kim2026infinipot}.
Once the memory reaches capacity, each new
observation requires a decision about which event to discard, and the effects
of these local decisions accumulate over time. Existing eviction policies
typically prioritize recency, novelty, coverage, or an estimate of the
importance of an individual event
~\citep{vitter1985random,aljundi2019gradient,zeng2026streamforest,duan2026think,dong2026objectstream,ge2026should}.
However, evaluations centered on retained information or downstream accuracy may leave unclear how successive evictions reshape the future supported by memory. The resulting drift can misalign long-horizon predictions with the unfolding
activity even under routine input variation, as illustrated in
Fig.~\ref{fig:teaser}(b). Yet preventing such drift by simply limiting replacement risks retaining
outdated history when the activity genuinely shifts. This creates a stability-plasticity
tension, also central to continual learning~\citep{rolnick2019experience}:
memory must resist drift under routine variation while remaining responsive
to genuine semantic change.

We argue that resolving this tension requires characterizing video memory
through the futures it supports. Different observed states may lead to
similar futures, whereas similar observed states may evolve into different
future activities. Following predictive representation learning
~\citep{oord2018representation,lecun2022path,assran2023self,
bardes2024revisiting}, we define the predictive state as the future
representations supported by the retained history. As shown in
Fig.~\ref{fig:teaser}(a), repeated local evictions can cause this state to
drift within a stable semantic regime. To resist such drift, predictive states should remain near a persistent
reference within a semantic regime, while allowing that reference to shift
when future semantics change. This motivates an adaptive predictive
attractor~\citep{hopfield1982neural,ramsauer2020hopfield,susman2019stable}. Because a
large departure may result either from an avoidable eviction or from a genuine
shift shared by all feasible updates, determining when to stabilize requires
comparing the counterfactual futures induced by alternative evictions.

To this end, we propose \textbf{SPACE}
(\textbf{S}parse \textbf{P}redictive \textbf{A}ttractor via
\textbf{C}ounterfactual \textbf{E}viction), a controller for fixed-capacity
streaming video memory. Rather than scoring stored events independently,
SPACE evaluates feasible evictions according to how they alter the future
supported by memory. Counterfactual predictions identify utility-preserving alternatives, while
the adaptive predictive attractor provides a reference for comparing their
predicted states and determining whether drift can be avoided without leaving
the current predictive regime. SPACE therefore
intervenes selectively to stabilize predictive dynamics when a useful,
state-consistent alternative exists and otherwise allows the memory to adapt.
As summarized in Fig.~\ref{fig:teaser}(c), this approach improves predictive
stability and long-horizon retention without sacrificing broader memory
capabilities.

To evaluate how well streaming memory preserves predictive utility while
remaining stable under routine variation and adaptive to genuine semantic
change, we introduce \textbf{MABS-Bench}, a controlled benchmark for
fixed-capacity memory under matched causal streams. It measures four
complementary properties: future-task sufficiency, within-regime predictive
stability, responsiveness to genuine semantic transitions, and recovery
toward a paired clean trajectory after controlled memory corruption.
Together, these dimensions help distinguish useful stability from mere inertia.

The main contributions of this paper are summarized as follows:
\begin{itemize}
\item We formulate fixed-capacity streaming video memory as a counterfactual control problem over transitions in a predictive state space and identify uncontrolled slot churn as a source of predictive-state drift under routine streaming variation.
\item We propose SPACE, a sparse eviction controller that combines JEPA-based counterfactual future utility with slow predictive-basin geometry to form an adaptive and recoverable predictive attractor through sparse, basin-aware interventions.
\item We introduce MABS-Bench, a matched-stream benchmark that jointly measures downstream utility, predictive-state dynamics, and recovery after controlled memory perturbations under a shared causal-stream protocol. Across multiple video datasets, SPACE improves the dataset-native task metric while reducing predictive-state drift.
\end{itemize}

\section{Related Work}
\paragraph{Streaming and long-form video modeling.}
Video models designed for long or continuous streams must preserve information
from an expanding history while keeping inference and storage costs bounded
~\citep{song2024moviechat,diko2025rewind,zhang2025flash,huang2025online}.
Recent systems address this constraint by organizing observations into
persistent event hierarchies~\citep{zeng2026streamforest,ge2026should},
maintaining object-centric anchors~\citep{dong2026objectstream}, compressing
multimodal key-value states~\citep{sun2026omnimem}, or constructing
historical event memories for long-horizon prediction and
control~\citep{wang2026dim}. These approaches demonstrate the value of
structured history for preserving task-relevant information under a bounded
memory budget.

\paragraph{Memory selection and eviction.}
Fixed-capacity memory requires a policy for deciding which observations to
retain. Classical strategies include FIFO and reservoir sampling
~\citep{vitter1985random}, while learned or training-free policies rank items
according to recency, novelty, redundancy, attention, or influence
~\citep{aljundi2019gradient,aljundi2019online}. More structured criteria
preserve complementary events, perturbation-sensitive states
~\citep{sun2026omnimem}, persistent objects~\citep{dong2026objectstream}, or
set-level coverage~\citep{duan2026think}. These methods therefore determine
what to retain based primarily on properties of individual items or of the
retained set.

\paragraph{Predictive representation learning and memory dynamics.}
Predictive coding and joint-embedding predictive architectures learn
representations by predicting future or target embeddings from observed
context rather than reconstructing pixels
~\citep{oord2018representation,lecun2022path,assran2023self}. V-JEPA extends
feature prediction to video and demonstrates that this objective captures both
appearance and motion semantics~\citep{bardes2024revisiting}. Separately, slow
feature analysis extracts temporally persistent coordinates by suppressing
rapid variation~\citep{6790128}. SPACE connects these ideas: a frozen
multi-horizon JEPA evaluates the futures supported by alternative memories,
while slow coordinates derived from its prediction trajectories define the
predictive state space used for counterfactual eviction control.

\section{Method}
\begin{figure}
    \centering
    \includegraphics[width=1\linewidth]{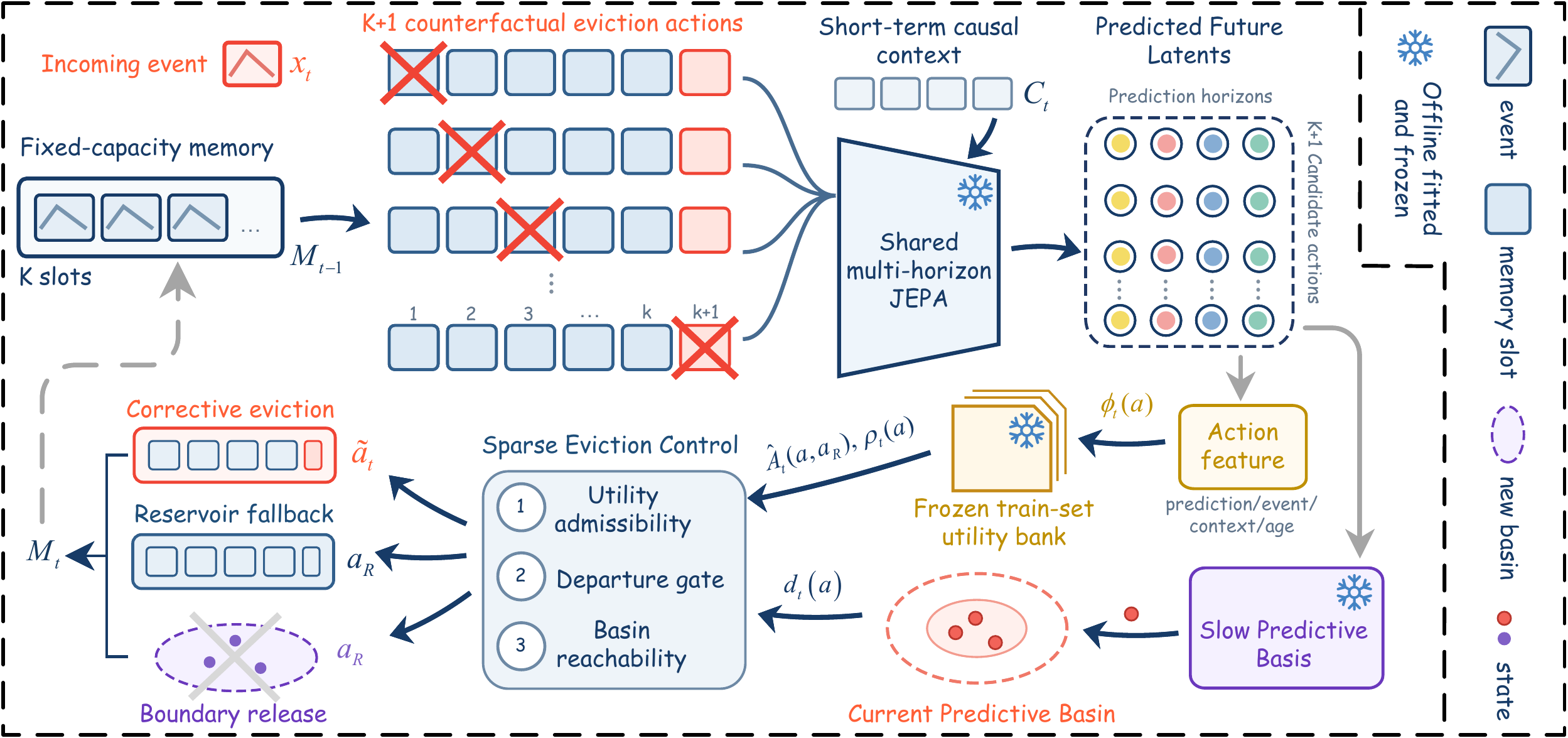}
    \caption{\textbf{SPACE overview.} Candidate eviction actions are evaluated
    by a frozen multi-horizon JEPA and selected using complementary
    future-utility and predictive-basin signals.}
    \label{fig:method}
\end{figure}

SPACE treats memory eviction as sparse control over the predictive state
supported by memory. Future utility alone may permit avoidable state drift,
whereas indiscriminately stabilizing the current state may prevent adaptation
to genuine regime change. SPACE uses Reservoir as the nominal transition and
intervenes only when a future-useful alternative can correct an avoidable
departure from the current predictive basin.
Figure~\ref{fig:method} summarizes the framework. At each full-memory update,
the incoming event and $K$ stored events define $K+1$ counterfactual eviction
actions. A shared frozen multi-horizon JEPA predicts the futures supported by
the resulting memories. These predictions feed complementary utility and
predictive-basin branches, which assess departure, utility admissibility, and
basin reachability. SPACE then chooses a corrective eviction or Reservoir
fallback, triggering boundary release when the previous basin becomes
unreachable. During deployment, all components remain frozen; only the memory,
basin prototype, and causal statistics evolve online.

\subsection{Memory as Predictive Dynamics}
\label{sec:predictive_dynamics}

The effect of removing an event depends on which other events remain. We
therefore evaluate each feasible post-eviction memory by the future
representations it supports. This makes the candidate-induced predictive
state, rather than an isolated stored event, the object of control in SPACE. A
streaming model maintains $K$ timestamped long-term events
$M_{t-1}=\{(m_i,\tau_i)\}_{i=1}^{K}$. Here, $t$ indexes memory updates,
$x_t$ is the event that has just left the short-term window, and the next
$L$ observed tokens form the more recent, non-overlapping causal context
$C_t=(c_{t,1},\ldots,c_{t,L})$. Events are inserted directly while free slots
remain. Once the memory is full, an action
$a\in\mathcal A_t=\{0,\ldots,K\}$ removes one item from
$M_{t-1}\cup\{x_t\}$, inducing
\begin{equation}
    M_t^{(a)}=\operatorname{Evict}(M_{t-1}\cup\{x_t\},a).
    \label{eq:counterfactual_memory}
\end{equation}
Thus, one action rejects $x_t$, while the other $K$ actions replace an
existing slot.

A shared multi-horizon JEPA predictor~\citep{assran2023self,bardes2024revisiting}
evaluates every candidate memory:
\begin{equation}
    \hat z_{t,h}^{(a)}=f_\theta(M_t^{(a)},C_t,h),
    \qquad h\in\mathcal H.
    \label{eq:counterfactual_prediction}
\end{equation}
Because all actions share $f_\theta$ and $C_t$, differences among their
predicted future latents isolate the effect of eviction. SPACE uses these
differences to estimate both future utility and movement in the predictive
state space; no ground-truth action labels or boundaries enter the controller.

\subsection{Predictive Attractor via Counterfactual Eviction}
\label{sec:predictive_attractor}

Future utility and predictive stability address different failure modes and
should not be optimized in isolation. Utility alone may favor an action that
preserves predictive accuracy but causes avoidable state departure, whereas
minimizing state change may make memory too inert to follow a genuine regime
shift. SPACE therefore combines the signals asymmetrically rather than as a
single score: future utility determines admissible alternatives, whereas
predictive-basin geometry determines whether intervention is necessary and
the current regime remains reachable. We denote the counter-based Reservoir
action by $a_R$ and use it as the nominal transition.

\paragraph{Counterfactual predictive utility.}
The utility branch first determines whether there is evidence that a candidate
preserves future predictions better than the nominal Reservoir action. On
training streams, the multi-horizon future-prediction cost of candidate $a$ is
\begin{equation}
    u_t(a)=\sum_{h\in\mathcal H}w_h
    \left[1-\cos(\hat z_{t,h}^{(a)},z_{t+h})\right],
    \label{eq:counterfactual_jepa_loss}
\end{equation}
where $w_h$ are normalized horizon weights. We retain only anchors whose
prediction and rollout targets all remain within the video. To account for
delayed consequences, we first execute $a$ and then follow Reservoir under the
same subsequent observations and random draws. The resulting rollout cost is
$q_t^{(R)}(a)=R^{-1}\sum_{j=0}^{R-1}u_{t+j}(a\!\rightarrow\!a_R)$.
The utility bank stores the standardized relative target
\begin{equation}
    \Delta C_t(a,a_R)=
    \beta\frac{u_t(a)-u_t(a_R)}{\sigma_{\mathrm{short}}}
    +(1-\beta)\frac{q_t^{(R)}(a)-q_t^{(R)}(a_R)}{\sigma_{\mathrm{roll}}},
    \label{eq:counterfactual_loss}
\end{equation}
where each scale is estimated from the training bank. Negative values favor
$a$ over Reservoir; SPACE uses the sign-reversed quantity as a positive
utility advantage. Computing this target requires future observations and is
therefore restricted to offline preparation. We store each training target
together with the corresponding action feature
\begin{equation}
    \phi_t(a)=\operatorname{concat}\!\left(
    \operatorname{vec}_{h}(W_\theta\hat z_{t,h}^{(a)}),
    W_\theta v_t^{(a)},W_\theta\bar c_t,\log(1+\gamma_t^{(a)})\right),
    \label{eq:action_feature}
\end{equation}
where $W_\theta$ is the frozen JEPA input projector,
$v_t^{(a)}$ and $\gamma_t^{(a)}$ denote the evicted event and its age,
respectively, and $\bar c_t$ is the mean context. We collect the bank by
enumerating actions along train-only FIFO and Reservoir trajectories. At
deployment, SPACE computes the same feature from the observed prefix and
retrieves similar training states and actions to estimate the counterfactual
cost, which is no longer observable online. The utility branch therefore
provides a support-aware admissibility test; it does not by itself select the
executed action.

\paragraph{Predictive basin dynamics.}
A utility-admissible action may still cause an unnecessary state transition,
whereas a large state change may instead reflect a genuine change in the
underlying predictive regime. Distinguishing these cases requires a temporal
reference that suppresses routine local variation while retaining persistent
predictive change. We therefore construct a second signal from the temporal
structure of the same JEPA predictions. Let
$y_{t,h}^{(a)}=\operatorname{norm}(W_\theta\hat z_{t,h}^{(a)})$, and write
$y_{t,h}$ for the corresponding prediction realized along a Reservoir
trajectory. From train-only Reservoir trajectories, we compute the
lag-$\delta$ displacement covariance
\begin{equation}
    \Sigma_\delta=\mathbb E_{p}\,\mathbb E_{t,h\mid p}
    \left[(y_{t,h}-y_{t-\delta,h})(y_{t,h}-y_{t-\delta,h})^\top\right],
    \label{eq:lagged_displacement_covariance}
\end{equation}
where participants $p$ receive equal weight. The $r$ leading generalized
eigenvectors of
\begin{equation}
    \Sigma_{\delta_l}v=\eta(\Sigma_{\delta_s}+\epsilon I)v,
    \qquad \delta_l>\delta_s,
    \label{eq:slow_generalized_eigenproblem}
\end{equation}
ordered by decreasing $\eta$, form $P_{\mathrm{slow}}$. These directions
exhibit low short-lag change relative to long-lag displacement, thereby
retaining persistent predictive evolution while suppressing routine local
variation. Applying the shared basis to every horizon yields the
action-conditioned predictive state
\begin{equation}
    s_t^{(a)}=\operatorname{norm}\!\left(
    \operatorname{vec}_{h}[P_{\mathrm{slow}}^\top y_{t,h}^{(a)}]\right).
    \label{eq:predictive_signature}
\end{equation}

The slow predictive state provides coordinates for the dynamics, but the
controller also requires a causal reference for the regime currently
supported by memory. SPACE represents this regime by a unit-norm basin
prototype $\mu_t$. For candidate $a$, its distance to the current basin is
$d_t(a)=1-\langle s_t^{(a)},\mu_{t-1}\rangle$; smaller values indicate a more
conservative transition. After executing $a_t$,
\begin{equation}
    \mu_t=\operatorname{norm}\!\left(
    (1-\alpha_t)\mu_{t-1}+\alpha_t s_t^{(a_t)}\right).
    \label{eq:basin_update}
\end{equation}
During normal operation, $\alpha_t=\alpha_{\mathrm{slow}}$, so the prototype
tracks only persistent predictive change. The Reservoir-selected state at the
first full-memory update initializes $\mu$. SPACE continues to follow Reservoir
during the calibration warm-up period while accumulating the causal histories
used by the controller. We write $\operatorname{RZ}(v;\mathcal D)$ for a
causal robust z-score computed from prior values in $\mathcal D$ using their
median and median absolute deviation.

\paragraph{Sparse eviction control.}

At test time, SPACE combines the two branches using utility-support, departure,
and reachability tests. A two-stage cosine retrieval first identifies similar
training states and then matches candidate actions within that neighborhood.
Let $\widehat A_t(a,a_R)$ be the retrieved mean advantage after subtracting
$\lambda$ times the combined standard error of $a$ and $a_R$. Retrieval
support $\rho_t(a)$ is the mean cosine similarity of the matched training
actions. SPACE admits only candidates for which both the candidate and the
Reservoir reference are sufficiently supported,
\begin{equation}
    \mathcal A_t^U=\{a_R\}\cup
    \{a:\widehat A_t(a,a_R)\geq\tau_U,\;
    \min(\rho_t(a),\rho_t(a_R))\geq\tau_S\}.
    \label{eq:utility_supported_set}
\end{equation}
Utility admissibility alone is not a sufficient reason to override Reservoir,
because ordinary updates may cause small and harmless predictive-state
changes. We therefore next test whether the nominal transition induces a
departure that another feasible action could have avoided. Across all $K+1$
actions, the restoration score
$r_t=[d_t(a_R)-\min_a d_t(a)]_+$ measures whether the default Reservoir
transition produces an avoidable departure. Let $\mathcal R_{t-1}$ contain the
preceding restoration scores. The corresponding causal gate is
$g_t^{\mathrm{dep}}=\sigma((\operatorname{RZ}(r_t;\mathcal R_{t-1})-
\tau_D)/T_D)$. The candidate considered for intervention is
$\tilde a_t=\arg\min_{a\in\mathcal A_t^U}d_t(a)$. Let
$\mathcal D_{t-1}$ contain the previously realized basin distances and define
$z_t^{\mathrm{reach}}=\operatorname{RZ}(d_t(\tilde a_t);\mathcal D_{t-1})$.
Both histories exclude the current update. SPACE selects
\begin{equation}
    a_t
    =
    \begin{cases}
        \tilde a_t,
        & g_t^{\mathrm{dep}}\geq\tau_G
          \ \land
          z_t^{\mathrm{reach}}\leq\tau_R,\\
        a_R,
        & \text{otherwise}.
    \end{cases}
    \label{eq:space_policy}
\end{equation}
This rule distinguishes three cases. When the departure gate is inactive,
SPACE retains the nominal Reservoir transition. When the gate is active and a
utility-admissible candidate remains reachable, SPACE executes the candidate
closest to the current basin. If $z_t^{\mathrm{reach}}>\tau_R$, SPACE follows
$a_R$ and triggers boundary release by setting
$\alpha_t=\alpha_{\mathrm{boundary}}$; otherwise,
$\alpha_t=\alpha_{\mathrm{slow}}$. Thus, SPACE corrects supported departures
within a reachable basin and rapidly relocates the prototype only when that
basin becomes unreachable.

\subsection{Learning and Deployment}
\label{sec:learning_deployment}

Offline preparation separates predictive representation learning from
controller construction. First, we train the causal Transformer JEPA on
sampled training-stream memory configurations using a weighted multi-horizon
cosine prediction loss. This stage optimizes $\theta$ using only
future-prediction targets and requires no counterfactual action supervision.
Second, after freezing the predictor, we replay train-only Reservoir
trajectories and fit $P_{\mathrm{slow}}$ from their projected prediction
dynamics. Third, at sampled full-memory states from train-only FIFO and
Reservoir trajectories, we enumerate all actions. For each candidate memory
$M_t^{(a)}$, the same prediction loss defines the action-conditioned cost
$u_t(a)$, from which we construct the counterfactual utility target and store
it together with the corresponding action feature in the utility bank.
At deployment, the encoder, JEPA, slow basis, and utility bank remain frozen.
Each full-memory update evaluates all $K+1$ candidates in a single JEPA batch,
retrieves their utility estimates, applies Eq.~\ref{eq:space_policy}, and
updates only the causal basin prototype and its histories through
Eq.~\ref{eq:basin_update}. Future observations are used only to construct
offline training targets; every deployment decision depends solely on the
observed prefix and causal statistics.

\section{Experiments}

\subsection{Implementation Details}
\label{sec:implementation_details}

Unless specified otherwise, all methods use a memory capacity of $K=16$ and
prediction horizons $\mathcal H=\{1,4,16,64\}$. SPACE uses a rank-$32$ slow
predictive basis with $(\delta_s,\delta_l)=(1,32)$, an $R=32$ rollout, and
equal standardized weights $\beta=0.5$ for the one-step and rollout costs. We
set $(\alpha_{\mathrm{slow}},\alpha_{\mathrm{boundary}})=(0.05,1)$ and
$(\tau_U,\tau_S,\tau_G,\tau_R)=(0.05,0.5,0.5,3.0)$. The structural settings were fixed in advance. Controller thresholds
were selected once on Assembly101 using the prespecified procedure in
Appendix~\ref{app:implementation_details} and were then fixed across all cross-dataset
experiments. Full details of predictor optimization, retrieval and calibration
settings, and hyperparameter sensitivity are provided in Appendix~\ref{app:implementation_details} and Appendix~\ref{app:sensitivity}.

\subsection{MABS-Bench}
\label{sec:mabs_bench}

Conventional downstream metrics measure what a fixed-capacity memory retains,
but not how its state evolves under continual updates or responds to
disruptions. We therefore introduce \textbf{MABS-Bench}
(\textbf{M}emory \textbf{A}daptation and \textbf{B}ehavioral
\textbf{S}tability Benchmark), which jointly evaluates future-task
sufficiency, within-regime predictive stability, transition responsiveness,
and perturbation recovery. Together, these dimensions characterize whether a
memory remains useful and stable under routine variation, adapts to genuine
semantic changes, and recovers from controlled disruptions.

\paragraph{Dataset and evaluation protocol.}
We conduct the main evaluation on
Assembly101~\citep{sener2022assembly101} using the participant-disjoint
validation split and 24-way multi-label verb prediction. Official
TSM~\citep{lin2019tsm} representations are sampled every 0.5 seconds, so the
prediction horizons $\mathcal H=\{1,4,16,64\}$ correspond to 0.5, 2, 8, and
32 seconds, respectively. The predictor, slow basis, utility bank, and
downstream action decoder are fitted on the training split. The same task--stability trend holds on Breakfast Actions~\citep{kuehne2014language}, IKEA ASM~\citep{ben2021ikea},
and EPIC-KITCHENS-100~\citep{damen2022rescaling} (Appendix~\ref{app:cross_dataset}).

All policies share the same causal stream, encoder, predictor, memory capacity,
and downstream evaluator. At time $t$, each receives the observed prefix,
current memory, and corresponding predictor outputs. Measurements are averaged
within recordings and equally weighted across participants. Gains are computed
against Reservoir on the same recording. The main table reports mean
$\pm$ sample standard deviation over three complete predictor-and-policy seeds.
Participant-level paired inference is detailed in
Appendix~\ref{app:statistical_protocol}.

To evaluate recovery from controlled disruptions, we intervene within an
Assembly101 semantic regime by replacing either $25\%$ or $50\%$ of the memory
slots with representations from an earlier, semantically different state in
the same video. The intervention preserves timestamps and slot metadata.
The clean and perturbed branches subsequently receive the same video stream,
and their predictive-state trajectories are compared at offsets
$h\in\{0,1,2,4,8,16\}$. Multiple interventions are first averaged within each
recording and then aggregated across participants.

\paragraph{Compared methods.}
We compare SPACE with standard eviction policies, predictor-guided policies,
recent memory-selection methods, and component variants of SPACE. Standard
policies include FIFO, Reservoir sampling~\citep{vitter1985random}, Surprise,
and Similarity, whereas predictor-guided baselines include Attention and
Self-Influence. We further adapt
StreamForest~\citep{zeng2026streamforest},
OmniMem~\citep{sun2026omnimem},
ObjectStream~\citep{dong2026objectstream}, and
NovaCov~\citep{duan2026think} to the shared MABS-Bench interface and
$K$-slot budget. Adaptation details and scope are provided in
Appendix~\ref{app:baseline_adapters}; these are controlled common-interface adaptations and are not intended to establish the relative performance of the original systems in their native settings. Utility-only applies counterfactual predictive utility
without basin geometry, whereas State-only applies departure and reachability
control without utility filtering.

\subsection{Evaluation Metrics}
\label{sec:evaluation_metrics}

\paragraph{Future-task sufficiency.}
We measure whether the retained memory supports future-action prediction.
\textbf{Action NLL} is the negative log-likelihood of the ground-truth future
labels under the action decoder applied to JEPA predictions. We average NLL
across the four prediction horizons and report horizon-specific gains at H1,
H4, H16, and H64. We additionally report
\textbf{mAP}, \textbf{LRAP}, and \textbf{Recall@5} for multi-label prediction.

\paragraph{Basin stability and boundary adaptation.}
Let $s_t$ denote the predictive state produced by a memory policy. For
Assembly101, let $q_t\in\{0,1\}^{24}$ denote the evaluation-only multi-hot
vector of annotated verbs active at time $t$. We restrict evaluation to
foreground regimes. Within-regime transitions retain the same nonempty
active-label set, whereas semantic boundaries comprise foreground onsets and
changes in the active-label set. Transitions from foreground to background are
excluded because background denotes the absence of an annotated action rather
than a coherent destination regime. \textbf{Basin drift} measures the expected
cosine distance between consecutive predictive states within the same
annotated regime:
\begin{equation}
D_{\mathrm{within}}
=
\mathbb E\!\left[
1-\cos(s_t,s_{t+1})
\mid q_t=q_{t+1}\neq\mathbf 0
\right].
\label{eq:within_basin_drift}
\end{equation}
Lower basin drift indicates greater within-regime stability. For adaptation, \textbf{boundary selectivity} compares predictive-state
displacement across semantic boundaries with within-regime drift:
\[
S_{\mathrm{boundary}}
=
\log\!\left(
\frac{D_{\mathrm{boundary}}+\epsilon}
     {D_{\mathrm{within}}+\epsilon}
\right),
\]
where $D_{\mathrm{boundary}}$ is the mean cosine displacement over foreground
onsets and foreground label-set changes as defined above. Higher selectivity
indicates that the predictive state changes more strongly at semantic
boundaries than within a regime.

\paragraph{Perturbation recovery.}
For a controlled memory perturbation, let $d_h$ denote the cosine distance
between the perturbed predictive state and the paired clean state after $h$
subsequent updates. \textbf{Predictive recovery} is
$\operatorname{Rec}(h)=(d_0-d_h)/\max(d_0,\epsilon)$, where
$\epsilon=10^{-8}$ is used only as a numerical safeguard. Larger values
indicate stronger contraction toward the clean trajectory. \textbf{Recovery
AUC} summarizes the trajectory as the normalized area under
$\operatorname{Rec}(h)$ across the evaluated offsets. We further report
\textbf{memory recovery} based on the convergence of the perturbed and clean
memory sets, and \textbf{task recovery} based on the convergence of their
future-action predictions.

\section{Results}

\subsection{Main Results on MABS-Bench}
\label{sec:main_results}

Table~\ref{tab:assembly101_main} shows that SPACE achieves the strongest
utility--stability trade-off among the compared methods. It achieves
the best or tied-best performance on every task metric, with its clearest
advantage at H64, while reducing within-regime drift relative to Reservoir in
all three seeds and achieving the highest boundary selectivity. The paired
gains are consistent across participants: Action NLL improves for 8 of 10
participants (95\% CI $[0.0010,0.0039]$, $p=0.0059$), and H64 improves for 9
of 10 participants ($p=0.0029$; Appendix~\ref{app:robustness}). Although
Attention and State-only achieve lower absolute drift, both degrade Action NLL
and H64 relative to Reservoir, indicating that minimizing state variation
alone does not preserve useful future information. 
An independent audit evaluator shows similar task and drift trends
under an independent predictive geometry (Appendix~\ref{app:independent_evaluator}).

\begin{table*}[t]
    \centering
    \caption{
        MABS-Bench results on Assembly101~\citep{sener2022assembly101} over
        three complete predictor-and-policy seeds. Values are mean $\pm$
        sample standard deviation. H64 gain is paired against Reservoir within
        the same seed. Best mean values are \textbf{bold} and second-best mean
        values are \underline{underlined}.
    }
    \label{tab:assembly101_main}
    \setlength{\tabcolsep}{3.4pt}
    \resizebox{\textwidth}{!}{%
    \begin{tabular}{lccccccc}
        \toprule
        Method
        & Action NLL $\downarrow$
        & mAP $\uparrow$
        & LRAP $\uparrow$
        & Recall@5 $\uparrow$
        & H64 gain $\uparrow$
        & Basin drift ($\times10^{-5}$) $\downarrow$
        & Boundary selectivity $\uparrow$ \\
        \midrule

        FIFO
        & $3.6820\pm0.0091$
        & $0.1339\pm0.0005$
        & $0.4743\pm0.0017$
        & $0.7030\pm0.0012$
        & $-0.0561\pm0.0246$
        & $8.7180\pm2.343$
        
        & $0.8229\pm0.0220$ \\

        Reservoir
        & $\underline{3.6315\pm0.0209}$
        & $\underline{0.1466\pm0.0012}$
        & $\underline{0.4875\pm0.0021}$
        & $\underline{0.7123\pm0.0026}$
        & $0.0000\pm0.0000$
        & $8.9420\pm2.429$
        & $0.8177\pm0.0206$ \\

        Surprise
        & $3.6633\pm0.0218$
        & $0.1443\pm0.0008$
        & $0.4797\pm0.0026$
        & $0.7080\pm0.0019$
        & $-0.0590\pm0.0075$
        & $9.3870\pm2.486$
        
        & $0.8135\pm0.0180$ \\

        Similarity
        & $3.6451\pm0.0201$
        & $0.1459\pm0.0010$
        & $0.4841\pm0.0022$
        & $0.7109\pm0.0020$
        & $-0.0179\pm0.0022$
        & $8.8410\pm2.488$
       
        & $0.8178\pm0.0232$ \\

        \midrule

        Attention
        & $3.6632\pm0.0264$
        & $0.1375\pm0.0022$
        & $0.4793\pm0.0019$
        & $0.7060\pm0.0012$
        & $-0.0296\pm0.0162$
        & $\mathbf{8.1840\pm2.344}$
        
        & $0.8254\pm0.0187$ \\

        Self-Influence
        & $3.6391\pm0.0218$
        & $\mathbf{0.1468\pm0.0013}$
        & $0.4870\pm0.0021$
        & $0.7115\pm0.0011$
        & $\underline{0.0033\pm0.0022}$
        & $8.3530\pm2.492$
        
        & $0.8155\pm0.0203$ \\

        \midrule

        StreamForest
        & $3.6557\pm0.0205$
        & $0.1446\pm0.0005$
        & $0.4814\pm0.0023$
        & $0.7093\pm0.0028$
        & $-0.0401\pm0.0034$
        & $9.1000\pm2.492$
        
        & $0.8247\pm0.0188$ \\

        OmniMem
        & $3.6398\pm0.0205$
        & $\mathbf{0.1468\pm0.0009}$
        & $0.4860\pm0.0023$
        & $0.7110\pm0.0024$
        & $-0.0161\pm0.0015$
        & $9.0510\pm2.460$
        
        & $0.8177\pm0.0233$ \\

        ObjectStream
        & $3.6657\pm0.0218$
        & $0.1438\pm0.0014$
        & $0.4798\pm0.0023$
        & $0.7084\pm0.0026$
        & $-0.0577\pm0.0036$
        & $9.1320\pm2.514$
        
        & $0.8208\pm0.0197$ \\

        NovaCov
        & $3.6610\pm0.0244$
        & $0.1376\pm0.0014$
        & $0.4783\pm0.0021$
        & $0.7073\pm0.0006$
        & $-0.0291\pm0.0167$
        & $8.6080\pm2.392$
        
        & $0.8214\pm0.0222$ \\

        \midrule

        Utility-only
        & $3.6403\pm0.0294$
        & $0.1437\pm0.0019$
        & $0.4858\pm0.0013$
        & $0.7119\pm0.0007$
        & $-0.0051\pm0.0227$
        & $8.8590\pm2.387$
        
        & $0.8123\pm0.0176$ \\

        State-only
        & $3.6522\pm0.0299$
        & $0.1401\pm0.0012$
        & $0.4822\pm0.0019$
        & $0.7094\pm0.0003$
        & $-0.0172\pm0.0224$
        & $\underline{8.3400\pm2.347}$
        
        & $\underline{0.8263\pm0.0172}$ \\

        \textbf{SPACE}
        & $\mathbf{3.6291\pm0.0209}$
        & $\mathbf{0.1468\pm0.0011}$
        & $\mathbf{0.4883\pm0.0018}$
        & $\mathbf{0.7131\pm0.0028}$
        & $\mathbf{0.0068\pm0.0016}$
        & $8.6900\pm2.421$
        
        & $\mathbf{0.8270\pm0.0219}$ \\

        \bottomrule
    \end{tabular}%
    }
\end{table*}

\subsection{Perturbation Recovery and Future Semantics}
\label{sec:attractor_recovery}

\paragraph{Recovery from controlled perturbations.}

\begin{table*}[t]
    \centering
    \caption{
        Recovery from controlled memory perturbations on Assembly101.
        Predictive recovery measures contraction toward the paired clean
        trajectory, and recovery AUC summarizes offsets H1--H16. Best values
        within each perturbation level are \textbf{bold} and second-best values
        are \underline{underlined}.
    }
    \label{tab:recovery}
    \setlength{\tabcolsep}{6pt}
    \resizebox{0.75\textwidth}{!}{%
    \begin{tabular}{llcccc}
        \toprule
        Corruption
        & Method
        & Predictive AUC $\uparrow$
        & H16 recovery $\uparrow$
        & Memory AUC $\uparrow$
        & Task AUC $\uparrow$ \\
        \midrule

        \multirow{4}{*}{$25\%$}
        & Reservoir
        & $\mathbf{0.0328}$
        & $\mathbf{0.1209}$
        & $0.0417$
        & $\underline{0.0399}$ \\

        & Utility-only
        & $-0.3954$
        & $-0.8692$
        & $0.0191$
        & $-0.2098$ \\

        & State-only
        & $-0.3740$
        & $-0.8425$
        & $\mathbf{0.0533}$
        & $-0.3587$ \\

        & \textbf{SPACE}
        & $\underline{0.0309}$
        & $\underline{0.0727}$
        & $\underline{0.0426}$
        & $\mathbf{0.0517}$ \\

        \midrule

        \multirow{4}{*}{$50\%$}
        & Reservoir
        & $\underline{0.0376}$
        & $\underline{0.0770}$
        & $\underline{0.0239}$
        & $\underline{0.0308}$ \\

        & Utility-only
        & $-0.4916$
        & $-0.7216$
        & $0.0082$
        & $-0.3305$ \\

        & State-only
        & $-0.2861$
        & $-0.7355$
        & $\mathbf{0.0534}$
        & $-0.7077$ \\

        & \textbf{SPACE}
        & $\mathbf{0.0525}$
        & $\mathbf{0.0881}$
        & $0.0218$
        & $\mathbf{0.0485}$ \\

        \bottomrule
    \end{tabular}%
    }
\end{table*}

Table~\ref{tab:recovery} evaluates recovery after replacing $25\%$ or $50\%$
of the memory slots with representations from earlier, semantically different
states. SPACE yields positive predictive, memory, and task recovery at both
corruption levels. At $25\%$, Reservoir achieves higher predictive AUC and H16
recovery, whereas SPACE achieves the highest task AUC. Because this milder
intervention produces smaller initial separations $d_0$, the normalized
ordering is more sensitive to the denominator; we therefore interpret this
result as evidence of positive recovery rather than as evidence that SPACE
outperforms Reservoir. At $50\%$, SPACE achieves the highest predictive AUC,
H16 recovery, and task AUC. The predictive advantage of SPACE also remains
positive under the minimum-$d_0$ sensitivity analysis in
Appendix~\ref{app:robustness}.

Figure~\ref{fig:attractor_recovery} shows the recovery trajectories under
$50\%$ corruption. Utility-only and State-only progressively diverge from
their paired clean trajectories, whereas SPACE maintains positive
contraction. This contrast supports combining counterfactual future utility
with predictive-basin control.

\begin{figure*}[t]
    \centering
    \includegraphics[width=\textwidth]{
        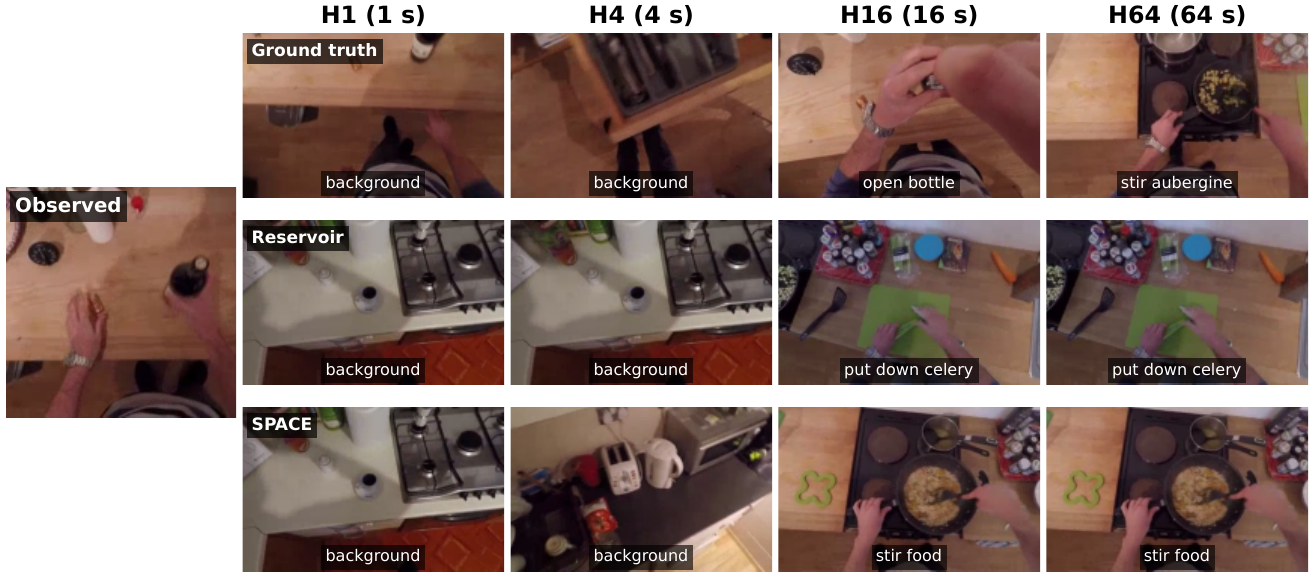
    }
    \caption{
        Future-semantic visualization on EPIC-KITCHENS-100. The left column
        shows the observed context, followed by predictions at H1, H4, H16,
        and H64. The rows show the ground-truth future and the nearest training
        frames retrieved using future representations predicted from Reservoir
        and SPACE memories.
    }
    \label{fig:epic_future_semantics}
\end{figure*}

\paragraph{Future-semantic visualization.}
Figure~\ref{fig:epic_future_semantics} visualizes the future representations
supported by memories produced by Reservoir and SPACE on
EPIC-KITCHENS-100~\citep{damen2022rescaling}. At H64, the ground-truth
activity is \emph{stir aubergine}. SPACE retrieves \emph{stir food},
preserving the long-term action semantics, whereas Reservoir retrieves
\emph{put down celery}. This semantic difference is consistent with the lower
H64 action-prediction loss of SPACE for the same observed context.

\begin{wrapfigure}[19]{r}{0.5\textwidth}
    \centering
    \includegraphics[width=\linewidth]{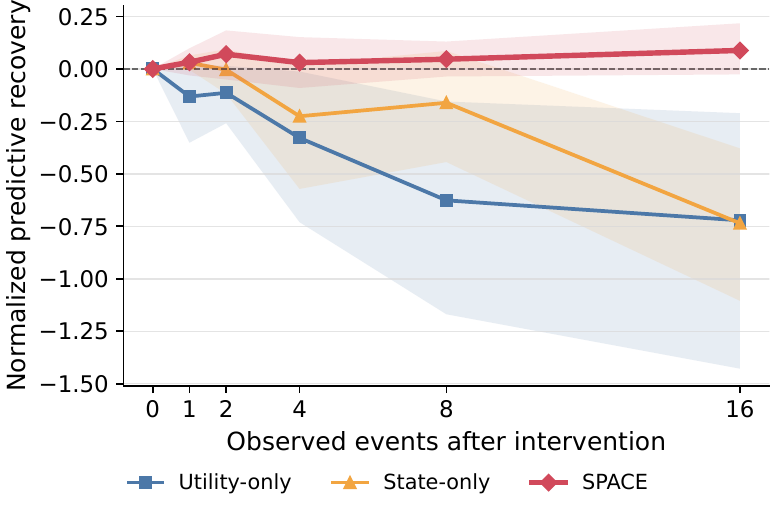}
    \caption{
    Predictive-state recovery after $50\%$ perturbations.
    Positive values denote contraction toward the paired clean trajectory;
    shading shows 95\% participant-bootstrap confidence intervals.
    }
    \label{fig:attractor_recovery}
\end{wrapfigure}

\subsection{Ablation Analysis}
\label{sec:ablation}

Table~\ref{tab:component_ablation} isolates the roles of predictive utility
and state control. Removing the utility gate produces the clearest failure
mode: task performance and H64 deteriorate despite substantially greater drift
reduction, indicating that contraction alone does not preserve useful future
information. The remaining ablations primarily redistribute this trade-off:
some improve an individual horizon or drift measure, but SPACE retains the
lowest mean Action NLL and the most balanced multi-horizon stability profile.
The comparison with the full-latent predictive state is also consistent with
the slow basis suppressing local variation without sacrificing aggregate
utility. SPACE is therefore designed to balance future utility and stability
rather than maximize contraction or performance at any single horizon.
Because several differences are small relative to their across-seed standard
deviations, we interpret fine-grained differences among variants
descriptively.

\begin{table*}[t]
    \centering
    \caption{
        Component ablation on Assembly101 over predictor-and-policy seeds.
        Within each seed, all variants share the same stream, predictor,
        utility bank, memory capacity, and downstream decoder. Horizon-specific
        NLL and drift gains are paired against Reservoir. Values are mean $\pm$
        sample standard deviation. Best mean values are \textbf{bold} and
        second-best mean values are \underline{underlined}.
    }
    \label{tab:component_ablation}
    \resizebox{\textwidth}{!}{%
    \begin{tabular}{lccccccc}
        \toprule
        Variant
        & Action NLL $\downarrow$
        & H1 gain ($\times10^{-2}$) $\uparrow$
        & H4 gain ($\times10^{-2}$) $\uparrow$
        & H16 gain ($\times10^{-2}$) $\uparrow$
        & H64 gain ($\times10^{-2}$) $\uparrow$
        & Drift gain ($\times10^{-6}$) $\uparrow$
        & Boundary selectivity $\uparrow$ \\
        \midrule

        \textbf{SPACE}
        & $\mathbf{3.6291\pm0.0209}$
        & $\mathbf{0.0189\pm0.0131}$
        & $\mathbf{-0.0007\pm0.0015}$
        & $\mathbf{0.2574\pm0.0029}$
        & \underline{$0.6844\pm0.1626$}
        & \underline{$2.5203\pm0.0804$}
        & $\mathbf{0.8270\pm0.0219}$ \\

        w/o Departure gate
        & $3.6434\pm0.0153$
        & $-2.1477\pm0.0318$
        & $-2.2982\pm1.6290$
        & $-1.7154\pm0.0896$
        & $\mathbf{1.4013\pm0.7037}$
        & $0.7356\pm0.1854$
        & $0.8249\pm0.0003$ \\

        w/o Utility gate
        & $3.6522\pm0.0299$
        & $-1.6018\pm0.5118$
        & $-2.6354\pm0.7932$
        & $-2.3207\pm0.2242$
        & $-1.7221\pm2.2370$
        & $\mathbf{6.0203\pm0.8612}$
        & $0.8263\pm0.0172$ \\

        w/o Reachability gate
        & \underline{$3.6314\pm0.0207$}
        & \underline{$-0.2293\pm0.0042$}
        & \underline{$-0.2801\pm0.0028$}
        & \underline{$0.0713\pm0.0120$}
        & $0.4780\pm0.4729$
        & $0.5623\pm0.0147$
        & $0.8199\pm0.0164$ \\

        Full-latent predictive state
        & $3.6431\pm0.0210$
        & $-1.4174\pm0.0020$
        & $-1.4993\pm0.0873$
        & $-1.1734\pm0.0199$
        & $-0.5559\pm0.0716$
        & $0.5138\pm0.0085$
        & $0.8261\pm0.0097$ \\

        w/o Boundary release
        & $3.6437\pm0.0209$
        & $-1.4370\pm0.0057$
        & $-1.5063\pm0.0991$
        & $-1.1458\pm0.0380$
        & $-0.7908\pm0.0203$
        & $0.3192\pm0.0084$
        & \underline{$0.8268\pm0.0172$} \\

        \bottomrule
    \end{tabular}%
    }
\end{table*}

\section{Conclusion}

We introduced SPACE, which treats fixed-capacity streaming memory eviction as
counterfactual control over future-predictive dynamics. By combining
counterfactual future utility with predictive-basin control, SPACE is designed
to preserve useful future information while limiting avoidable predictive-state
drift and remaining responsive to genuine regime changes. Together with
MABS-Bench, our experiments show that, among the compared policies, this
formulation provides a better balance between downstream utility and
predictive-state dynamics, while the resulting memory dynamics remain
recoverable after controlled perturbations. More broadly, these results
support evaluating streaming memory not only by what it retains, but also by
how its predictive state evolves under continual updates. SPACE currently
relies on a predictor and utility bank trained offline and frozen during
deployment; adapting these components under larger distribution shifts remains
an important direction for future work on adaptive streaming memory.

\bibliography{iclr2027_conference}
\bibliographystyle{iclr2027_conference}

\clearpage
\appendix
\section{Implementation and Reproducibility Details}
\label{app:implementation_details}

This appendix provides the full implementation configuration,
dataset-specific evaluation protocols, statistical aggregation and inference
procedures, hyperparameter sensitivity analyses, and robustness audits.

\subsection{Complete Configuration and Selection Protocol}

Table~\ref{tab:full_configuration} summarizes the default configuration.
Structural choices, including the prediction horizons, slow-basis construction,
and rollout definition, were fixed in advance. Controller thresholds were
selected once on the Assembly101 validation split using a prespecified
feasibility-first protocol and were then frozen for all cross-dataset
experiments. Feasibility was assessed before performance ranking, and no
weighted composite score was used. The slow basis and utility bank were
constructed exclusively from training streams.

\begin{table*}[h]
\centering
\caption{
Default implementation configuration. Unless stated otherwise, the same
settings are used across datasets.
}
\label{tab:full_configuration}
\small
\setlength{\tabcolsep}{5pt}
\begin{tabular}{@{}lll@{}}
\toprule
Component & Parameter & Value \\
\midrule
\multirow{11}{*}{JEPA predictor}
& Hidden dimension & $512$ \\
& Transformer layers / heads & $4 / 8$ \\
& Feed-forward dimension & $2048$ \\
& Recent-context length & $8$ \\
& Prediction horizons $\mathcal H$ & $\{1,4,16,64\}$ \\
& Horizon weights $w_h$ & uniform ($1/4$) \\
& Dropout / memory dropout & $0.1 / 0.15$ \\
& Training epochs / batch size & $30 / 128$ \\
& AdamW learning rate / weight decay & $3\times10^{-4} / 0.05$ \\
& Learning-rate schedule / warmup & cosine / $5\%$ linear \\
& Numerical precision & bfloat16 \\
\midrule
\multirow{5}{*}{Slow basis}
& Rank $r$ & $32$ \\
& Short / long lag $(\delta_s,\delta_l)$ & $(1,32)$ \\
& Covariance regularization & $10^{-3}$ \\
& Participant weighting & equal \\
& Basis sharing across horizons & yes \\
\midrule
\multirow{6}{*}{Utility retrieval}
& Memory capacity $K$ & $16$ \\
& Rollout length $R$ & $32$ \\
& One-step / rollout weight $\beta$ & $0.5$ \\
& State / action neighbors & $32 / 32$ \\
& Retrieval similarity & cosine \\
& Uncertainty multiplier $\lambda$ & $1.0$ \\
\midrule
\multirow{6}{*}{Controller}
& Prototype rates $(\alpha_{\mathrm{slow}},
  \alpha_{\mathrm{boundary}})$ & $(0.05,1.0)$ \\
& Utility / support thresholds $(\tau_U,\tau_S)$ & $(0.05,0.5)$ \\
& Departure threshold / temperature $(\tau_D,T_D)$ & $(1.0,0.5)$ \\
& Restoration / reachability thresholds $(\tau_G,\tau_R)$ & $(0.5,3.0)$ \\
& History length / warmup & $32 / 16$ \\
& Minimum robust scale & $10^{-5}$ \\
\bottomrule
\end{tabular}
\end{table*}



\subsection{Statistical Aggregation and Inference}
\label{app:statistical_protocol}

For each seed, measurements are first averaged within each recording and then
across recordings within each participant. Participants receive equal weight
regardless of the number of recordings. A seed-level estimate is obtained by
averaging the participant-level values, and the main tables report the mean
and sample standard deviation of these estimates across three seeds.

For paired inference, each method and Reservoir are evaluated on the same
recordings. Their recording-level measurements are first differenced within
each seed, then averaged within each participant, and finally averaged across
seeds. We resample the 10 participants with replacement for 10,000 bootstrap
replicates and report the 2.5th and 97.5th percentiles as the 95\% confidence
interval. We additionally apply an exact one-sided sign-flip test to the 10
participant-level paired gains, with the alternative that their mean is
greater than zero. Because controller thresholds are selected on the Assembly101 validation
split, these intervals and tests are reported as post-selection measures of
participant-level consistency rather than as fully held-out confirmatory
inference.

\subsection{Adaptation of Baselines}
\label{app:baseline_adapters}

We implement literature-inspired memory adapters under the shared
MABS-Bench interface. All adapters process the same frozen visual features
and causal stream, maintain at most $K$ long-term vectors, and are evaluated
with the same JEPA predictor and downstream decoder. Incoming events fill
empty slots until capacity is reached. Thereafter, OmniMem and NovaCov select
$K$ vectors from the $K+1$ available candidates, whereas StreamForest and
ObjectStream may either merge an incoming event into an existing slot or
replace a stored vector. Thus, all methods share the same slot budget, although
merging adapters may retain aggregated representations rather than individual
observed events.

\paragraph{StreamForest-style event merging.}
We approximate persistent event memory with a flat collection of $K$
event representatives. The incoming event is matched to the representative
with the highest cosine similarity weighted by exponential recency decay.
We use a temporal scale of 64 updates and merge when this affinity is at
least $0.75$. Otherwise, the representative with the lowest weighted
combination of uniqueness, recency, and persistence is replaced, using
weights $(0.5,0.25,0.25)$. Uniqueness is one minus the maximum cosine
similarity to any other stored representative, and persistence is the
logarithm of one plus the accumulated count. This adapter does not reconstruct
the original hierarchical event forest.

\paragraph{OmniMem-style perturbation-aware selection.}
We enumerate all $K+1$ single-item evictions and obtain their multi-horizon
forecasts from the shared frozen JEPA. Each candidate is scored by its mean
horizon-wise cosine distance from the average candidate forecast, plus $0.1$
times the retained-set redundancy. Redundancy is the mean, over retained
vectors, of the maximum cosine similarity to another retained vector. The
candidate with the lowest score is selected. This adapter uses a visual-only
forecast-distortion surrogate rather than the original attention-output
distortion criterion, audio--visual budget allocation, or KV-cache
compression.

\paragraph{ObjectStream-style persistent anchors.}
We treat stored vectors as latent event anchors. An incoming event is merged
into its most similar anchor when their cosine similarity is at least $0.85$.
Otherwise, it replaces the anchor with the lowest retention score, which
combines persistence, distinctiveness, and recency with weights
$(0.5,0.3,0.2)$. Distinctiveness uses the same nearest-neighbor uniqueness
measure as above, and recency has a half-life of 64 updates. These anchors
operate on event-level features; the adapter does not perform patch-level
latent-object discovery or maintain object-conditioned temporal residuals.

For both merging adapters, an anchor with count $n$ is updated with
coefficient $\min(0.25,1/(n+1))$. Its timestamp is refreshed to that of the
incoming event, and its count is incremented; new anchors start with count
one.

\paragraph{NovaCov-style set coverage.}
For each possible eviction, we score the retained set using weighted
facility-location coverage over the $K+1$ candidate vectors and recent
context. Each reference contributes its maximum cosine similarity to a
retained vector. Candidate-reference weights decay with an age half-life
of 64 updates, whereas recent-context references receive unit weight.
We select the eviction that maximizes total coverage. This adapter retains
a set-based coverage criterion but does not implement the original separate
historical reference bank or dual-branch novelty-aware objective.

These comparisons evaluate the specified adapters under a common
representation, causal stream, and memory budget. They should not be
interpreted as establishing the relative performance of the original systems
in their native settings.

\section{Capacity and Hyperparameter Sensitivity}
\label{app:sensitivity}

Table~\ref{tab:capacity_sensitivity} reports a single-seed diagnostic across
memory capacities $K\in\{8,16,32\}$ and one-factor variations in the utility,
departure, and reachability thresholds. SPACE retains positive NLL, H16, H64,
and drift gains at every tested capacity and threshold setting. The default
configuration yields the largest NLL, H16, and drift gains, whereas several
alternatives yield larger H64 gains. We therefore interpret these results as
evidence of local robustness rather than universal hyperparameter optimality.

\begin{table}[t]
    \centering
    \caption{
Single-seed capacity and one-factor threshold sensitivity on Assembly101.
Gains are paired against Reservoir with the same memory capacity. Best values
within each block are bold.
}
    \label{tab:capacity_sensitivity}
    \resizebox{\textwidth}{!}{%
    \begin{tabular}{lcccc}
        \toprule
        Setting
        & $\Delta$NLL $\uparrow$
        & H16 gain$\uparrow$
        & H64 gain$\uparrow$
        & Drift gain $\uparrow$ \\
        \midrule
        $K=8$
        & $0.000818$
        & $0.000811$
        & $0.002413$
        & $8.833{\times}10^{-8}$ \\

        $K=16$
        & $\mathbf{0.002387}$
        & $\mathbf{0.002552}$
        & $\mathbf{0.007412}$
        & $\mathbf{2.5101{\times}10^{-6}}$ \\

        $K=32$
        & $0.001384$
        & $0.001633$
        & $0.004260$
        & $4.303{\times}10^{-7}$ \\
        \midrule
        Default $(\tau_U,\tau_G,\tau_R)=(0.05,0.5,3.0)$
        & $\mathbf{0.002387}$
        & $\mathbf{0.002552}$
        & $0.007412$
        & $\mathbf{2.5101{\times}10^{-6}}$ \\
        
        $\tau_U=0.00$
        & $0.002239$
        & $0.002413$
        & $\mathbf{0.008183}$
        & $7.153{\times}10^{-7}$ \\

        $\tau_U=0.10$
        & $0.001963$
        & $0.001522$
        & $0.006591$
        & $4.254{\times}10^{-7}$ \\

        $\tau_G=0.25$
        & $0.002247$
        & $0.002218$
        & $0.007858$
        & $6.721{\times}10^{-7}$ \\

        $\tau_G=0.75$
        & $0.001334$
        & $0.000459$
        & $0.005246$
        & $3.049{\times}10^{-7}$ \\

        $\tau_R=2.0$
        & $0.001683$
        & $0.001544$
        & $0.005759$
        & $4.622{\times}10^{-7}$ \\

        $\tau_R=4.0$
        & $0.001850$
        & $0.001646$
        & $0.007291$
        & $4.625{\times}10^{-7}$ \\
        \bottomrule
    \end{tabular}%
    }
\end{table}

\section{Robustness and Statistical Audits}
\label{app:robustness}

\subsection{Participant-Level Consistency}

Table~\ref{tab:participant_inference} evaluates whether the gains of SPACE over
Reservoir are consistent across Assembly101 validation participants. Within
each seed, recording-level paired gains are averaged within each participant;
the resulting participant-level gains are then averaged across the three
seeds. All metrics are oriented so that positive values favor SPACE. Action
NLL and H16 gains are positive for 8 of 10 participants, the H64 gain is
positive for 9 of 10, and the basin-drift gain is positive for all 10.
All participant-bootstrap confidence intervals exclude zero. Equal
participant weighting prevents participants with more recordings from
dominating the estimates, while the positive counts show that the gains are
broadly distributed across participants.

\begin{table}[t]
\centering
\caption{
Participant-level paired inference for SPACE relative to Reservoir on
Assembly101. Gains are averaged across seeds within each participant, and
positive values favor SPACE.
}
\label{tab:participant_inference}
\small
\begin{tabular}{lcccc}
\toprule
Metric & Mean gain & Positive & 95\% bootstrap CI & $p$ \\
\midrule
Action NLL
& $0.002400$ & $8/10$ & $[0.00105,0.00392]$ & $0.0059$ \\
H16 NLL
& $0.002574$ & $8/10$ & $[0.00115,0.00422]$ & $0.0059$ \\
H64 NLL
& $0.006844$ & $9/10$ & $[0.00296,0.01190]$ & $0.0029$ \\
Basin drift
& $2.5203\times10^{-6}$ & $10/10$
& $[1.97{\times}10^{-6},3.56{\times}10^{-6}]$ & $0.0010$ \\
\bottomrule
\end{tabular}
\end{table}

\subsection{Recovery Normalization Audit}

Predictive recovery is normalized by the initial distance $d_0$, so very small
initial separations could amplify small absolute changes.
Table~\ref{tab:initial_recovery_distance} reports the intervention-level
distribution of $d_0$ before recording- and participant-level aggregation.
The analysis includes 62 valid interventions per policy at $25\%$ corruption
and 59 at $50\%$ corruption. The numerical safeguard
$\epsilon=10^{-8}$ was never active because the minimum observed distance was
$8.34\times10^{-7}$. Stronger corruption generally produces larger initial
separations. For SPACE, the median $d_0$ increases from
$1.37\times10^{-5}$ at $25\%$ corruption to $3.25\times10^{-5}$ at
$50\%$, while the fraction below $10^{-5}$ decreases from $35.5\%$ to
$5.1\%$.

We additionally repeat the paired SPACE--Reservoir comparison after retaining
only interventions for which both methods exceed the same minimum-$d_0$
threshold. At $50\%$ corruption, the SPACE-minus-Reservoir recovery-AUC gain
remains positive at thresholds
${2\times10^{-6},5\times10^{-6},10^{-5}}$, with corresponding gains of
${0.0145,0.0276,0.0322}$, compared with $0.0150$ without filtering. At
$25\%$ corruption, the paired gain changes sign across thresholds. We
therefore interpret the milder setting as evidence of absolute contraction
rather than a robust improvement over Reservoir.
\begin{table}[t]
\centering
\caption{
Distribution of the intervention-level initial predictive distance $d_0$.
Intervals denote the 5th--95th percentiles.
}
\resizebox{\textwidth}{!}{%
\label{tab:initial_recovery_distance}
\small
\begin{tabular}{llccc}
\toprule
Corruption & Method
& Median [5th, 95th]
& Minimum
& Fraction with $d_0<10^{-5}$ \\
\midrule
25\% & Reservoir
& $1.36{\times}10^{-5}$ [$2.15{\times}10^{-6}$, $6.73{\times}10^{-5}$]
& $1.07{\times}10^{-6}$ & 35.5\% \\
25\% & Utility-only
& $1.22{\times}10^{-5}$ [$2.40{\times}10^{-6}$, $7.78{\times}10^{-5}$]
& $1.01{\times}10^{-6}$ & 33.9\% \\
25\% & State-only
& $9.30{\times}10^{-6}$ [$1.14{\times}10^{-6}$, $6.81{\times}10^{-5}$]
& $8.94{\times}10^{-7}$ & 51.6\% \\
25\% & SPACE
& $1.37{\times}10^{-5}$ [$2.39{\times}10^{-6}$, $7.48{\times}10^{-5}$]
& $1.07{\times}10^{-6}$ & 35.5\% \\
\midrule
50\% & Reservoir
& $2.74{\times}10^{-5}$ [$3.22{\times}10^{-6}$, $9.89{\times}10^{-5}$]
& $1.67{\times}10^{-6}$ & 13.6\% \\
50\% & Utility-only
& $2.36{\times}10^{-5}$ [$2.82{\times}10^{-6}$, $9.56{\times}10^{-5}$]
& $1.61{\times}10^{-6}$ & 22.0\% \\
50\% & State-only
& $1.60{\times}10^{-5}$ [$2.95{\times}10^{-6}$, $7.88{\times}10^{-5}$]
& $8.34{\times}10^{-7}$ & 35.6\% \\
50\% & SPACE
& $3.25{\times}10^{-5}$ [$1.03{\times}10^{-5}$, $9.96{\times}10^{-5}$]
& $3.16{\times}10^{-6}$ & 5.1\% \\
\bottomrule
\end{tabular}
}
\end{table}

\section{Cross-Dataset Results}
\label{app:cross_dataset}

We extend the matched causal-stream evaluation to Breakfast Actions,
IKEA ASM, and EPIC-KITCHENS-100. Within each dataset, all policies
process the same observed streams with a shared memory capacity,
frozen predictor, and frozen decoder. Online decisions never access
future frames or action labels. These comparisons isolate the effect
of changing the memory policy within the fixed evaluation pipeline.

The tables report the same task and predictive-dynamics measurements
as Table~\ref{tab:assembly101_main}. For the single-label Breakfast
and IKEA tasks, Top-1 and Top-5 accuracy replace the multi-label mAP,
LRAP, and Recall@5 metrics used for Assembly101 and EPIC-KITCHENS-100.
H64 gain is the paired reduction in H64 Action NLL relative to
Reservoir.

Measurements are first averaged within each recording and then within
each participant, with equal weight assigned to participants. Gains
are computed relative to Reservoir on the same recording before
aggregation. Across the evaluated configurations, SPACE improves task
performance relative to Reservoir while reducing within-regime
predictive drift. These participant-balanced point estimates provide
complementary cross-dataset evidence, but do not quantify uncertainty
from participant sampling or training randomness.

\paragraph{Breakfast Actions.}
Table~\ref{tab:breakfast_results} shows that SPACE achieves the highest Top-1
and Top-5 accuracy, the lowest basin drift, and the highest boundary
selectivity. It also improves Action NLL and H64 relative to Reservoir,
although Attention retains the lowest Action NLL and the largest H64 gain.
These results indicate a favorable trade-off between future-task
utility and predictive stability under the evaluated setting.

\begin{table*}[t]
\centering
\caption{
Results on Breakfast Actions under the matched causal-stream protocol.
H64 gain is paired against Reservoir. Best mean values are \textbf{bold} and
second-best mean values are \underline{underlined}.
}
\label{tab:breakfast_results}
\setlength{\tabcolsep}{5pt}
\resizebox{\textwidth}{!}{%
\begin{tabular}{lcccccc}
\toprule
Method
& Action NLL $\downarrow$
& Top-1 $\uparrow$
& Top-5 $\uparrow$
& H64 gain $\uparrow$
& Basin drift ($\times10^{-3}$) $\downarrow$
& Boundary selectivity $\uparrow$ \\
\midrule
FIFO
& $2.6320$ & $0.3151$ & $0.6256$ & $0.0057$
& $\underline{2.7040}$ & $\underline{0.3538}$ \\
Reservoir
& $2.6474$ & $0.3173$ & $0.6152$ & $0.0000$
& $3.2163$ & $0.2991$ \\
Surprise
& $2.7044$ & $0.3046$ & $0.6007$ & $-0.1049$
& $3.6405$ & $0.3293$ \\
Similarity
& $2.6827$ & $0.3063$ & $0.6049$ & $-0.0686$
& $3.3748$ & $0.3335$ \\
\midrule
Attention
& $\mathbf{2.6126}$ & $0.3212$ & \underline{$0.6305$}
& $\mathbf{0.0571}$ & $2.8566$ & $0.3160$ \\
Self-Influence
& $2.6332$ & $0.3145$ & $0.6261$
& $\underline{0.0342}$ & $2.7257$ & $0.3426$ \\
\midrule
Utility-only
& $2.6445$ & $0.3185$ & $0.6141$ & $0.0173$
& $3.1083$ & $0.3233$ \\
State-only
& $2.6423$ & \underline{$0.3221$} & $0.6147$ & $0.0140$
& $2.9754$ & $0.3285$ \\
\textbf{SPACE}
& $\underline{2.6198}$ & $\mathbf{0.3299}$ & $\mathbf{0.6353}$ & $0.0322$
& $\mathbf{2.5013}$ & $\mathbf{0.3730}$ \\
\bottomrule
\end{tabular}%
}
\end{table*}

\paragraph{IKEA ASM.}
On IKEA ASM, SPACE achieves the lowest Action NLL, highest Top-1 accuracy,
largest H64 gain, lowest basin drift, and highest boundary selectivity, while
ranking second in Top-5 accuracy (Table~\ref{tab:ikea_results}). These results
show that the task and predictive-dynamics benefits of SPACE transfer to a
distinct single-label streaming regime rather than being specific to
Assembly101.

\begin{table*}[t]
\centering
\caption{
Results on the IKEA ASM sealed test split under the matched causal-stream
protocol. H64 gain is paired against Reservoir. Best mean values are
\textbf{bold} and second-best mean values are \underline{underlined}.
}
\label{tab:ikea_results}
\setlength{\tabcolsep}{5pt}
\resizebox{\textwidth}{!}{%
\begin{tabular}{lcccccc}
\toprule
Method
& Action NLL $\downarrow$
& Top-1 $\uparrow$
& Top-5 $\uparrow$
& H64 gain $\uparrow$
& Basin drift ($\times10^{-3}$) $\downarrow$
& Boundary selectivity $\uparrow$ \\
\midrule
FIFO
& $2.5686$ & $0.3176$ &$\mathbf{0.6489}$ & $0.0062$ & $1.0576$ & $0.1914$ \\
Reservoir
& $2.5786$ & $0.3146$ & $0.6482$ & $0.0000$
& $1.0782$ & $\underline{0.2718}$ \\
Surprise
& $2.6211$ & $0.2883$ & $0.6332$ & $-0.0591$
& $1.1446$ & $0.2654$ \\
Similarity
& $2.6365$ & $0.2686$ & $0.6269$ & $-0.0696$
& $1.0874$ & $0.2688$ \\
\midrule
Attention
& $\underline{2.5206}$ & $\underline{0.3506}$ & $0.6412$
& $\underline{0.0549}$ & $1.0085$ & $0.2133$ \\
Self-Influence
& $2.5947$ & $0.3078$ & $0.6445$ & $-0.0341$
& $\underline{0.9963}$ & $0.2439$ \\
\midrule
Utility-only
& $2.5626$ & $0.3226$ & $0.6451$ & $0.0049$
& $1.0825$ & $0.2666$ \\
State-only
& $2.5604$ & $0.3274$ & $0.6466$ & $0.0082$
& $1.0263$ & $0.2516$ \\
\textbf{SPACE}
& $\mathbf{2.4889}$ & $\mathbf{0.3736}$ & $\underline{0.6484}$
& $\mathbf{0.0873}$
& $\mathbf{0.9600}$ & $\mathbf{0.2730}$ \\
\bottomrule
\end{tabular}%
}
\end{table*}

\paragraph{EPIC-KITCHENS-100.}
Table~\ref{tab:epic_results} further confirms this pattern on the larger and
more diverse EPIC-KITCHENS-100 benchmark. SPACE achieves the lowest Action
NLL, highest LRAP, largest H64 gain, and lowest basin drift, while ranking
second in mAP, Recall@5, and boundary selectivity. SPACE improves future-task utility and within-regime stability
relative to Reservoir, while individual baselines remain stronger
on some metrics.

\begin{table*}[t]
\centering
\caption{
Results on EPIC-KITCHENS-100 under the matched causal-stream protocol. H64
gain is paired against Reservoir. Best mean values are \textbf{bold} and
second-best mean values are \underline{underlined}.
}
\label{tab:epic_results}
\setlength{\tabcolsep}{3.5pt}
\resizebox{\textwidth}{!}{%
\begin{tabular}{lccccccc}
\toprule
Method
& Action NLL $\downarrow$
& mAP $\uparrow$
& LRAP $\uparrow$
& Recall@5 $\uparrow$
& H64 gain $\uparrow$
& Basin drift ($\times10^{-3}$) $\downarrow$
& Boundary selectivity $\uparrow$ \\
\midrule
FIFO
& $4.3424$ & $0.2715$ & $\underline{0.4478}$ & $0.6198$
& $0.0125$ & $8.5610$ & $0.6106$ \\
Reservoir
& $4.3469$ & $0.2749$ & $0.4473$ & $0.6247$
& $0.0000$ & $9.2249$ & $0.5950$ \\
Surprise
& $4.3844$ & $0.2753$ & $0.4426$ & $0.6172$
& $-0.0879$ & $10.6138$ & $0.5888$ \\
Similarity
& $4.3832$ & $0.2746$ & $0.4432$ & $0.6147$
& $-0.0821$ & $9.6320$ & $0.5975$ \\
\midrule
Attention
& $4.3448$ & $0.2714$ & $0.4467$
& $\mathbf{0.6261}$ & $0.0162$
& $\underline{7.4006}$ & $\mathbf{0.6337}$ \\
Self-Influence
& $4.3519$ & $0.2720$ & $0.4458$ & $0.6228$
& $0.0037$ & $7.5461$ & $0.6257$ \\
\midrule
StreamForest
& $4.3615$ & $0.2733$ & $0.4454$ & $0.6202$
& $-0.0337$ & $9.0522$ & $0.6016$ \\
OmniMem
& $4.3635$ & $0.2742$ & $0.4448$ & $0.6187$
& $-0.0366$ & $9.8807$ & $0.5871$ \\
ObjectStream
& $4.3673$ & $0.2738$ & $0.4431$ & $0.6181$
& $-0.0414$ & $9.3291$ & $0.5942$ \\
NovaCov
& $4.3619$ & $\mathbf{0.2756}$ & $0.4463$ & $0.6182$
& $-0.0368$ & $8.5487$ & $0.5993$ \\
\midrule
Utility-only
& $\underline{4.3415}$ & $0.2748$ & $0.4454$ & $0.6241$
& $\underline{0.0182}$ & $8.8777$ & $0.6041$ \\
State-only
& $4.3459$ & $0.2743$ & $0.4467$ & $0.6248$
& $0.0082$ & $8.3987$ & $0.6175$ \\
\textbf{SPACE}
& $\mathbf{4.3245}$ & $\underline{0.2755}$ & $\mathbf{0.4530}$ & $\underline{0.6254}$
& $\mathbf{0.0433}$ & $\mathbf{7.3193}$ & $\underline{0.6289}$ \\
\bottomrule
\end{tabular}%
}
\end{table*}

\section{Runtime Cost and Memory Footprint}
\label{app:runtime_cost}

We profile the deployment-time costs of representative policies on
Assembly101 with memory capacity $K=16$. Measurements are obtained on an
NVIDIA A800 80 GB GPU.
Disk I/O, downstream metric computation, and trajectory logging are excluded
from the timed region. Wall-clock measurements are synchronized with the GPU
before and after each update.

For each of three predictor-and-policy seeds, we profile 15 recordings from
five validation participants. For each recording, we measure one full-memory
update after the 16-update controller calibration period, so that SPACE is
profiled under its normal deployment-time decision rule rather than during
calibration. Each measurement uses three untimed CUDA warm-up repetitions
followed by ten timed repetitions. Repetitions are first averaged within each
recording, recordings are then averaged within each participant, and
participants receive equal weight. The reported standard deviation is computed
over the five participant-level means after averaging across seeds.

\begin{table*}[t]
\centering
\caption{
Deployment-time cost on Assembly101 with $K=16$.
Wall time is the participant-balanced mean $\pm$ sample standard deviation
per full-memory update. ``Candidates'' counts candidate-equivalent
evaluations, whereas ``calls'' counts predictor forward passes after
batching. $\times$Res. is computed relative to Reservoir within each matched participant and seed before aggregation; it therefore need not equal the ratio of the displayed mean wall times. $\Delta$GPU denotes the peak memory allocated during the update
above the loaded pre-update state.
}
\label{tab:runtime_cost}
\resizebox{\textwidth}{!}{%
\small
\begin{tabular}{lrrrrrr}
\toprule
Method
& Candidates
& Calls
& Wall time (ms) $\downarrow$
& $\times$Res.
& $\Delta$GPU (MB) $\downarrow$
& Peak GPU (MB) $\downarrow$ \\
\midrule
FIFO
& 0 & 0 & $1.686 \pm 0.003$ & 1.67
& 0.78 & 925.68 \\
Reservoir
& 0 & 0 & $1.006 \pm 0.006$ & 1.00
& 0.65 & 925.55 \\
Attention
& 2 & 2 & $9.866 \pm 1.337$ & 9.89
& 4.28 & 931.89 \\
Self-Influence
& 18 & 18 & $61.698 \pm 0.068$ & 63.05
& 4.63 & 932.24 \\
Utility-only
& 17 & 1 & $8.950 \pm 0.042$ & 9.15
& 828.82 & 1756.43 \\
State-only
& 17 & 1 & $5.367 \pm 0.188$ & 5.47
& 12.63 & 940.24 \\
\textbf{SPACE}
& 17 & 1 & $9.870 \pm 0.038$ & 10.08
& 828.82 & 1756.43 \\
\bottomrule
\end{tabular}
}
\end{table*}

Table~\ref{tab:runtime_cost} summarizes the deployment-time latency
and memory footprint of the evaluated policies. At each full-memory update, SPACE evaluates all $K+1=17$ feasible actions in
one batched predictor pass. It requires $9.87$ ms per update, approximately
$10.1\times$ the cost of Reservoir but comparable to Attention and
$6.25\times$ faster than Self-Influence, which performs 18 separate predictor
calls. At the 0.5-second sampling interval used on Assembly101, the measured
SPACE update occupies approximately $2\%$ of the interval. The State-only and
Utility-only variants require $5.37$ ms and $8.95$ ms, respectively,
indicating the additional cost of combining the two branches.

SPACE reaches a peak allocated GPU footprint of approximately $1.76$ GB,
including $828.82$ MB of update-time allocation. The utility bank occupies
approximately $1.47$ GB of CPU memory and $740.24$ MB of GPU memory in the
profiled implementation. Absolute peak values include resident allocations
from the common benchmarking process and therefore should not be interpreted
as the standalone memory requirements of baselines that do not use the
utility bank.

Batching reduces the practical cost by replacing $K+1$ separate predictor
calls with one forward pass, but it does not remove the underlying
$\mathcal{O}(K)$ candidate computation or activation-memory scaling.
The measurements therefore establish that the overhead is practical at the
evaluated capacity $K=16$, rather than claiming cost independent of memory
capacity.

\section{Retrospective Utility Validation}
\label{app:utility_validation}

We retrospectively evaluate whether the train-only utility bank identifies
beneficial candidate evictions. Future observations used to construct these retrospective oracle targets are
accessed only after online decisions are fixed; they are not used to train the
utility bank or alter the evaluated online decisions. Oracle advantage is
measured relative to the same Reservoir action using the train-standardized
combination of one-step and $R=32$ rollout costs, with positive values
indicating lower future cost.

\begin{table}[t]
\centering
\caption{
Retrospective utility validation on Assembly101. Estimator metrics use 960
full-memory states with 17 candidate actions, and intervention metrics use
209 SPACE overrides with complete oracle targets. Hit@1 and Hit@3 are compared
with uniform ranking. Combined gain is reported in train-standardized cost
units.
}
\label{tab:utility_validation}
\small
\begin{tabular}{lcc}
\toprule
Metric & Observed & Reference \\
\midrule
\multicolumn{3}{l}{\emph{Estimator ordering}} \\
Pairwise sign accuracy $\uparrow$
& $61.98\%$ & $51.29\%$ majority \\
Hit@1 $\uparrow$
& $13.53\%$ & $5.88\%$ random \\
Hit@3 $\uparrow$
& $28.29\%$ & $17.65\%$ random \\
\midrule
\multicolumn{3}{l}{\emph{Executed interventions}} \\
Positive-gain overrides $\uparrow$
& $58.85\%$ & $50.00\%$ parity \\
Mean combined oracle gain $\uparrow$
& $+0.4166$ & $0$ (Reservoir) \\
\bottomrule
\end{tabular}
\end{table}

Table~\ref{tab:utility_validation} shows that the estimator exceeds the
majority sign baseline by $10.69$ percentage points. Its Hit@1 and Hit@3 rates
are approximately $2.30\times$ and $1.60\times$ their corresponding
uniform-ranking levels. The utility bank therefore provides informative
candidate screening rather than a precise ranking over all 17 actions,
consistent with its role as an admissibility filter rather than the final
action selector.

SPACE overrides Reservoir on only $3.44\%$ of full-memory updates. Among the
209 overrides with valid oracle targets, $58.85\%$ reduce retrospective future
cost. Although not every intervention improves utility in hindsight, the mean
combined oracle advantage remains positive at $+0.4166$, indicating that
beneficial corrections outweigh detrimental interventions on average. This
result is consistent with the asymmetric role of utility in SPACE: utility
screens supported alternatives, while basin geometry determines whether and
when to intervene.

\section{Independent-Evaluator Analysis}
\label{app:independent_evaluator}

The primary stability metrics are computed from predictions of the JEPA model
that also supports the SPACE controller. To test whether the observed behavior
is specific to this predictive geometry, we construct a separate audit
evaluator with a different architecture and disjoint parameters. The audit
predictor and its action decoder are trained only on the Assembly101 training
split and remain frozen throughout evaluation. Neither component accesses the
utility bank, slow predictive basis, basin prototype, controller thresholds,
or SPACE decisions.

We replay the frozen policy trajectories through this evaluator without
altering their eviction actions. The independent predictive state is formed
directly from the multi-horizon outputs of the audit predictor, without
applying $P_{\mathrm{slow}}$, and is used to recompute within-regime drift and
boundary selectivity. The accompanying action decoder provides an independent
measurement of future-task utility.

\begin{table*}[t]
\centering
\caption{
Evaluation with an independently trained audit predictor on Assembly101.
Values are participant-balanced means $\pm$ sample standard deviations across
the ten participant-level means after averaging policy seeds. Gains are paired
with Reservoir on the same recording. Best values are \textbf{bold}, and
second-best values are \underline{underlined}.
}
\resizebox{\textwidth}{!}{%
\label{tab:independent_evaluator}
\small
\begin{tabular}{lccccc}
\toprule
Method
& Audit Action NLL $\downarrow$
& H16 gain ($\times10^{-4}$) $\uparrow$
& H64 gain ($\times10^{-4}$) $\uparrow$
& Audit drift gain ($\times10^{-6}$) $\uparrow$
& Boundary selectivity $\uparrow$ \\
\midrule
Reservoir
& $\underline{3.5763\pm0.1073}$
& $0.0000\pm0.0000$
& $0.0000\pm0.0000$
& $0.0000\pm0.0000$
& $\underline{0.8568\pm0.0864}$ \\

Utility-only
& $3.5768\pm0.1082$
& $\underline{2.5895\pm10.7770}$
& $5.0917\pm10.9680$
& $2.0749\pm1.1209$
& $0.8564\pm0.0868$ \\

State-only
& $3.5777\pm0.1081$
& $-0.1472\pm9.1888$
& $\underline{5.2427\pm11.9840}$
& $\mathbf{22.2150\pm14.2020}$
& $0.8563\pm0.0881$ \\

\textbf{SPACE}
& $\mathbf{3.5761\pm0.1070}$
& $\mathbf{4.6237\pm4.6086}$
& $\mathbf{6.0321\pm4.5149}$
& $\underline{3.0900\pm4.5634}$
& $\mathbf{0.8569\pm0.0872}$ \\
\bottomrule
\end{tabular}
}
\end{table*}

As shown in Table~\ref{tab:independent_evaluator}, SPACE retains the lowest
Audit Action NLL and the largest positive gains at both H16 and H64 under the
independent evaluator. It also reduces within-regime drift while maintaining
boundary selectivity under the independent predictive geometry. State-only
produces substantially greater drift reduction but weaker future-task utility,
consistent with the main finding that stronger state contraction alone does not improve predictive utility.

Although the differences under the audit evaluator are small, their consistent
direction across future-task and state-dynamics metrics suggests that the
behavior of SPACE is not specific to the JEPA geometry used by the controller.
Because both evaluators operate on the same frozen TSM input representation,
we interpret these results as robustness to an independently trained evaluator
with a different architecture and disjoint parameters, rather than as complete
representation independence.

\section{Stability Beyond the Slow Predictive Space}
\label{app:projection_audit}

To assess whether the observed drift reduction depends on
$P_{\mathrm{slow}}$, we re-evaluate fixed policy trajectories using the
original JEPA predictor across three evaluation spaces. In addition to the
slow space, we consider the pre-slow space, which retains the JEPA input
projection but omits $P_{\mathrm{slow}}$, and the raw JEPA output space,
which omits both projections. The predictor and eviction decisions remain
unchanged; only the representation used to measure drift varies. 

\begin{table}[t]
\centering
\caption{
SPACE drift gains beyond the slow predictive space.
Gain is defined as
$D_{\mathrm{within}}^{\mathrm{Reservoir}}
-D_{\mathrm{within}}^{\mathrm{SPACE}}$;
positive values indicate lower within-regime drift.
Means and 95\% participant-level confidence intervals are reported in
units of $10^{-6}$. Positive counts indicate participants with positive
paired gains.
}
\label{tab:projection_audit}
\small
\begin{tabular}{lccc}
\toprule
Evaluation space
& Mean drift gain $\uparrow$
& 95\% CI
& Positive \\
\midrule
Slow
& $0.4958$
& $[0.2818,\,0.7513]$
& $5/5$ \\
Pre-slow
& $3.6922$
& $[2.4701,\,5.5429]$
& $5/5$ \\
Raw JEPA output
& $7.8954$
& $[5.5802,\,11.5720]$
& $5/5$ \\
\bottomrule
\end{tabular}
\end{table}

Table~\ref{tab:projection_audit}  shows positive drift gains across all three
spaces, with confidence intervals excluding zero and positive gains for all
five evaluated participants. In particular, the pre-slow result directly
isolates the effect of removing $P_{\mathrm{slow}}$, while the raw-output
result provides further evidence that the drift reduction is not confined to
the slow predictive geometry used by the controller. Gain magnitudes are not
directly comparable across representation spaces. Given the limited
participant coverage, we interpret this audit as complementary evidence of
stability beyond the slow projection rather than as evidence of
representation-independent stability.

\end{document}